\documentclass[11pt,a4paper]{article}
\usepackage[hyperref]{acl2019}
\usepackage{times}
\usepackage{latexsym}

\usepackage{url}
\hypersetup{breaklinks=true}

\aclfinalcopy
\def\aclpaperid{779}

\usepackage{comment, algorithm, amsmath, amsfonts, amsthm}
\usepackage{graphicx}
\usepackage{algpseudocode}
\algrenewcommand\algorithmicrequire{\textbf{Input:}}
\algrenewcommand\algorithmicensure{\textbf{Output:}}
\newcommand{\argmax}{\mathop{\rm argmax}\limits}
\newcommand{\argmin}{\mathop{\rm argmin}\limits}
\usepackage{color}

\theoremstyle{definition}
\newtheorem{defi}{Def.}

\title{Answering while Summarizing: Multi-task Learning  \\ for Multi-hop QA with Evidence Extraction}

\author{
		Kosuke Nishida$^1$,
		Kyosuke Nishida$^1$,
		Masaaki Nagata$^2$,\\
		\textbf{Atsushi Otsuka}$^1$,
		\textbf{Itsumi Saito}$^1$,
		\textbf{Hisako Asano}$^1$,
		\textbf{Junji Tomita}$^1$\\
		$^1$ \rm NTT Media Intelligence Laboratories, NTT Corporation\\
		$^2$ \rm NTT Communication Science Laboratories, NTT Corporation\\
		\tt kosuke.nishida.ap@hco.ntt.co.jp
}
\date{}

\begin{document}
	\maketitle
	\begin{abstract}
	Question answering (QA) using textual sources for purposes such as reading comprehension (RC) has attracted much attention. This study focuses on the task of \textit{explainable multi-hop QA}, which requires the system to return the answer with evidence sentences by reasoning and gathering disjoint pieces of the reference texts. It proposes the \textit{Query Focused Extractor} (QFE) model for evidence extraction and uses multi-task learning with the QA model. QFE is inspired by extractive summarization models; compared with the existing method, which extracts each evidence sentence independently, it sequentially extracts evidence sentences by using an RNN with an attention mechanism on the question sentence. It enables QFE to consider the dependency among the evidence sentences and cover important information in the question sentence. Experimental results show that QFE with a simple RC baseline model achieves a state-of-the-art evidence extraction score on HotpotQA. Although designed for RC, it also achieves a state-of-the-art evidence extraction score on FEVER, which is a recognizing textual entailment task on a large textual database.
	\end{abstract}
	
	\section{Introduction}
	Reading comprehension (RC) is a task that uses textual sources to answer any question. It has seen significant progress since the publication of numerous datasets such as SQuAD \cite{squad}.
	To achieve the goal of RC, systems must be able to reason over disjoint pieces of information in the reference texts. 
	Recently, multi-hop question answering (QA) datasets focusing on this capability, such as QAngaroo \cite{qangaroo} and HotpotQA \cite{hotpot}, have been released.

    Multi-hop QA faces two challenges. 
    The first is the difficulty of reasoning.
    It is difficult for the system to find the disjoint pieces of information as evidence and reason using the multiple pieces of such evidence. The second challenge is interpretability. The evidence used to reason is not necessarily located close to the answer, so it is difficult for users to verify the answer.

\begin{figure}[t!]
		\begin{center}
			\includegraphics[width =75mm]{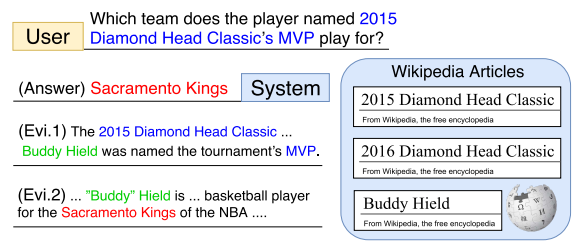} 
			\caption{Concept of explainable multi-hop QA.
			Given a question and multiple textual sources, the system extracts evidence sentences from the sources and returns the answer and the evidence.}
			\label{sample} 
		\end{center}	
	\end{figure}

	\citet{hotpot} released HotpotQA, an explainable multi-hop QA dataset, as shown in Figure \ref{sample}.
	Hotpot QA provides the evidence sentences of the answer for supervised learning.
	The evidence extraction in multi-hop QA is more difficult than that in other QA problems because the question itself may not provide a clue for finding evidence sentences.
	As shown in Figure \ref{sample}, the system finds an evidence sentence (Evidence 2) by relying on another evidence sentence (Evidence 1).
	The capability of being able to explicitly extract evidence is an advance towards meeting the above two challenges.

    Here, we propose a Query Focused Extractor (QFE) that is based on a summarization model.
    We regard the evidence extraction of the explainable multi-hop QA as a query-focused summarization task. Query-focused summarization is the task of summarizing the source document with regard to the given query.
	QFE sequentially extracts the evidence sentences by using an RNN with an attention mechanism on the question sentence, while the existing method extracts each evidence sentence independently.
	This query-aware recurrent structure enables QFE to consider the dependency among the evidence sentences and cover the important information in the question sentence.
    Our overall model uses multi-task learning with a QA model for answer selection and QFE for evidence extraction. The multi-task learning with QFE is general in the sense that it can be combined with any QA model.

    Moreover, we find that the recognizing textual entailment (RTE) task on a large textual database, FEVER \cite{fever}, can be regarded as an explainable multi-hop QA task. We confirm that QFE effectively extracts the evidence both on HotpotQA for RC and on FEVER for RTE.
    
	Our main contributions are as follows.
	\begin{itemize}
		\item We propose QFE for explainable multi-hop QA.
		We use the multi-task learning of the QA model for answer selection and QFE for evidence extraction.
		\item QFE adaptively determines the number of evidence  sentences by considering the dependency among the evidence sentences and the coverage of the question.
		\item QFE achieves state-of-the-art performance on both HotpotQA and FEVER in terms of the evidence extraction score and comparable performance to competitive models in terms of the answer selection score.
		QFE is the first model that outperformed the baseline on HotpotQA.
	\end{itemize}
	
	\section{Task Definition}
	\label{sec:Definition}
    Here, we re-define explainable multi-hop QA so that it includes the RC and the RTE tasks.
    \begin{defi} Explainable Multi-hop QA
		\begin{description}
			\item[\textbf{Input:}] Context $C$ (multiple texts), Query $Q$ (text)
			\item[\textbf{Output:}] Answer Type $A_T$ (label), Answer String $A_S$ (text), Evidence $E$ (multiple texts)
		\end{description}
		The \textbf{Context $C$} is regarded as one connected text in the model.
	    If the connected $C$ is too long (e.g.~over 2000 words), it is truncated.
		The \textbf{Query $Q$} is the query. The model answers $Q$ with an answer type $A_T$ or an answer string $A_S$. The \textbf{Answer Type $A_T$} is selected from the answer candidates, such as `Yes'. The answer candidates depend on the task setting. The \textbf{Answer String $A_S$} exists only if there are not enough answer candidates to answer $Q$. The answer string $A_S$ is a short span in $C$. \textbf{Evidence} $E$ consists of the sentences in $C$ and is required to answer $Q$.
	\end{defi}

    For RC, we tackle HotpotQA. In HotpotQA, the answer candidates are `Yes', `No', and `Span'. The answer string $A_S$ exists if and only if the answer type $A_T$ is `Span'. $C$ consists of ten Wikipedia paragraphs. The evidence $E$ consists of two or more sentences in $C$.

	For RTE, we tackle FEVER. In FEVER, the answer candidates are `Supports', `Refutes', and `Not Enough Info'. The answer string $A_S$ does not exist. $C$ is the Wikipedia database. 
	The evidence $E$ consists of the sentences in $C$. 

	\section{Proposed Method}
	
	This section first explains the overall model architecture, which contains our model as a module, and then the details of our QFE.
	\label{sec:proposed}
	
	\begin{figure}[t]
		\begin{center}
			\includegraphics[width =75mm]{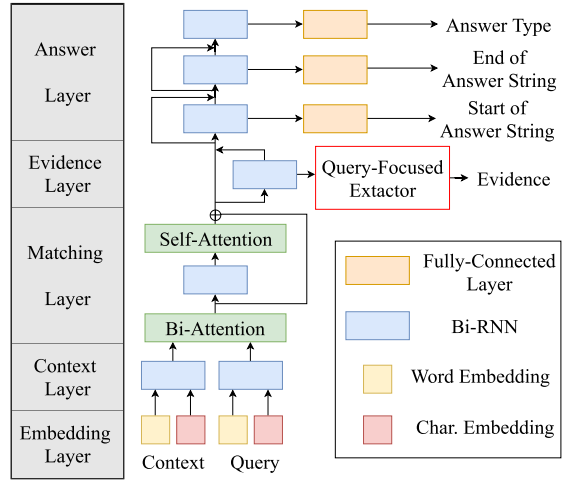} 
			\caption{Overall model architecture. The answer layer is the version for the RC task.}
			\label{hotpotmodel} 
		\end{center}	
	\end{figure}
	
	\subsection{Model Architecture}
	Except for the evidence layer, our model is the same as the baseline \cite{simple} used in HotpotQA \cite{hotpot}. Figure \ref{hotpotmodel} shows the model architecture. 
	The input of the model is the context $C$ and the query $Q$.
	The model has the following layers.
	
    \paragraph{The Word Embedding Layer} encodes $C$ and $Q$ 
	as sequences of word vectors.
	A word vector is the concatenation of a pre-trained word embedding and
	a character-based embedding obtained using a CNN \cite{char}.
	The outputs are $C_1 \in \mathbb{R}^{l_w \times d_w}, Q_1 \in \mathbb{R}^{m_w \times d_w}$,
	where $l_w$ is the length (in words) of $C$, $m_w$ is the length of $Q$ and
	$d_w$ is the size of the word vector.

	\paragraph{The Context Layer}
	encodes $C_1, Q_1$ as contextual vectors
	$C_2 \in \mathbb{R}^{l_w \times 2d_c}, Q_2 \in \mathbb{R}^{m_w \times 2d_c}$
	by using a bi-directional RNN (Bi-RNN),
	where $d_c$ is the output size of a uni-directional RNN.

	\paragraph{The Matching Layer}encodes $C_2, Q_2$ as matching vectors $C_3 \in \mathbb{R}^{l_w \times d_c}$
	by using bi-directional attention \cite{bidaf}, a Bi-RNN, and self-attention \cite{self}.
	
	\paragraph{The Evidence Layer} first encodes $C_3$
	as 
	$[\overrightarrow{C_4}; \overleftarrow{C_4}] \in \mathbb{R}^{l_w \times 2d_c}$ by a Bi-RNN.
	Let $j_1(i)$ be the index of the first word of the $i$-th sentence in $C$ and
	$j_2(i)$ be the index of the last word.
	We define the vector of the $i$-th sentence as:
	\[
	x_i =[\overrightarrow{c_{4, j_2(i)}}; \overleftarrow{c_{4, j_1(i)}} ]
	\in \mathbb{R}^{2d_c}.
	\]
	Here, $X \in \mathbb{R}^{l_s \times 2d_c}$ is the sentence-level context vectors,
	where $l_s$ is the number of sentences of $C$.
	
	QFE, described later, receives sentence-level context vectors
	$X\in \mathbb{R}^{l_s \times 2d_c}$ and
	the contextual query vectors $Q_2 \in \mathbb{R}^{m_w \times 2d_c}$ as Y.
	QFE outputs the probability distribution that
	the $i$-th sentence is the evidence:
	\begin{align}
	\Pr(i) = \mathrm{QFE}(X, Y=Q_2).
	\label{eq:qfe}
	\end{align}
	
	Then, the evidence layer concatenates
	the word-level vectors and the sentence-level vectors:
	\[
	c_{5,j} = [c_{3,j}; x_{i(j)}] \in \mathbb{R}^{3d_c},
	\]
	where the $j$-th word in $C$ is included in the $i(j)$-th sentence in $C$. 

	\paragraph{The Answer Layer} predicts the answer type $A_T$ and the answer string $A_S$
	from $C_5$.
	The layer has stacked Bi-RNNs.
	The output of each Bi-RNN is mapped to the probability distribution
	by the fully connected layer and the softmax function.
	
	For RC, the layer has three stacked Bi-RNNs.
	Each probability indicates the start of the answer string,
	$\hat{A}_{S1} \in \mathbb{R}^{l_w}$,
	the end of the answer string
	$\hat{A}_{S2} \in \mathbb{R}^{l_w}$,
	and the answer type,
	$\hat{A}_T \in \mathbb{R}^3$.
	For RTE, the layer has one Bi-RNN.
	The probability indicates the answer type.
	
	\paragraph{Loss Function:}
	Our model uses multi-task learning with a loss function
	$L = L_A + L_E$,
	where $L_A$ is the loss of the answer and 
	$L_E$ is the loss of the evidence.
	The answer loss 
	$L_A$ is the sum of the cross-entropy losses for
	all probability distributions obtained by the answer layer.
	The evidence loss $L_E$ is defined in subsection \ref{ssec:learning}.
	
	\subsection{Query Focused Extractor}
	\begin{figure}[t]
		\begin{center}
			\includegraphics[height =75mm, angle =270]{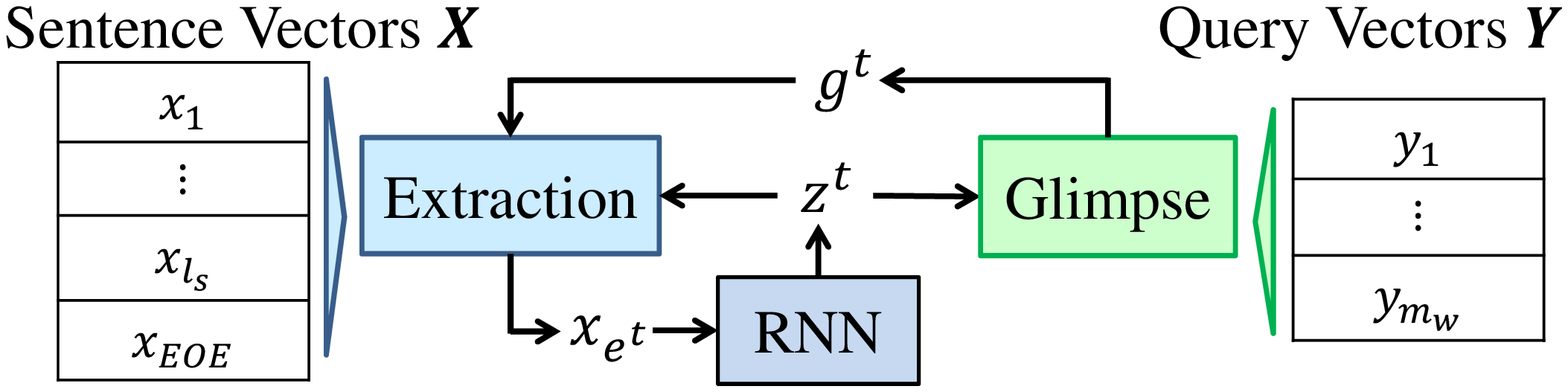} 
			\caption{Overview of Query Focused Extractor at step $t$. $z^t$ is the current summarization vector. $g^t$ is the query vector considering the current summarization. $e^t$ is the extracted sentence. $x_{e^t}$ updates the RNN state.
		}
			\label{proposedmodel}
		\end{center}	
	\end{figure}
    
    Query Focused Extractor (QFE) is shown as the red box in Figure \ref{hotpotmodel}. 
    QFE is an extension of the extractive summarization model of \citet{fast}, which is not for query-focused settings.
	Chen and Bansal used an attention mechanism to extract sentences from the source document such that the summary would cover the important information in the source document.
	To focus on the query, 
	QFE extracts sentences from $C$ with attention on $Q$
	such that the evidence covers the important information with respect to $Q$.
	Figure \ref{proposedmodel} shows an overview of QFE.
	
	The inputs of QFE are the sentence-level context vectors $X\in \mathbb{R}^{l_s \times 2d_c}$
	and contextual query vectors $Y\in \mathbb{R}^{m_w \times 2d_c}$.
    We define the timestep to be the operation to extract a sentence. QFE updates the state of the RNN (the dark blue box in Figure \ref{proposedmodel}) as follows:
	\[
	z^t =\textrm{RNN}(z^{t-1}, x_{e^t}) \in \mathbb{R}^{2d_c},
	\]
	where $e^t \in \{1,\cdots, l_s\}$ is the index of the sentence extracted at step $t$.
	We define  $E^t =\{e^1, \cdots , e^t\}$ to be the set of sentences extracted until step $t$.
	
	QFE extracts the $i$-th sentence according to the probability distribution (the light blue box):
	\[
	\Pr(i; E^{t-1}) =\textrm{softmax}_i(u^t_i)
	\]
	\[
	u^t_i =
	\begin{cases}
	v_p^\top \textrm{tanh}(W_{p1}x_i +W_{p2}g^t + W_{p3}z^t)\\
	\hspace{10.5eM} (i \not \in E^{t-1})\\
	-\infty	\hspace{8.5eM}(\textrm{otherwise})
	\end{cases}.
	\]
	Then, QFE selects $e^t =\argmax~\Pr (i; E^{t-1})$.
	
	Let $g^t$ be a query vector considering the importance at step $t$.
	We define $g^t$ as the glimpse vector \cite{glimpse} (the green box):
	\[
	\begin{split}
	g^t &=\sum_j \alpha^t_j W_{g1}y_j \in \mathbb{R}^{2d_c} \\
	\alpha^t &=\textrm{softmax}(a^t) \in \mathbb{R}^{m_w} \\
	a^t_j &=v_g^\top \textrm{tanh}(W_{g1}y_j +W_{g2}z^t).
	\end{split}
	\]

	The initial state of the RNN is the vector obtained via the fully connected layer and the max pooling from $X$. 
	All parameters $W_{\cdot} \in \mathbb{R}^{2d_c \times 2d_c}$ and $v_{\cdot} \in \mathbb{R}^{2d_c}$ are trainable.
	
	\subsection{Training Phase}
	\label{ssec:learning}
	In the training phase, we use teacher-forcing to make the loss function.
	The loss of the evidence $L_E$ is the negative log likelihood 
	regularized by a coverage mechanism \cite{pointer}:
	\[\begin{split}
	L_E =-\sum_{t=1}^{|E|}&\log \left(\max_{i \in E\setminus E^{t-1}} \Pr(i; E^{t-1})\right) \\
	&+\sum_i \min(c^t_i, \alpha^t_i).
	\end{split}\]
	The max operation in the first term enables the sentence with the highest probability to be extracted.
	This operation means that QFE extracts the sentences in the predicted importance order.
	On the other hand, the evidence does not have the ground truth order in which it is to be extracted, so the loss function ignores the order of the evidence sentences.
	The coverage vector $c^t$ is defined as
	$
	c^t =\sum_{\tau=1}^{t-1} \alpha^\tau.
	$
	
	In order to learn the terminal condition of the extraction, QFE adds a dummy sentence, called the EOE sentence, to the sentence set. When the EOE sentence is extracted, QFE terminates the extraction. The EOE sentence vector $x_{EOE}\in \mathbb{R}^{2d_c}$ is a trainable parameter in the model, so $x_{EOE}$ is independent of the samples. We train the model to extract the EOE sentence after all evidence.
	
	\subsection{Test Phase}
	\label{ssec:test}
	In the test phase, QFE terminates the extraction by reaching the EOE sentence. 
	The predicted evidence is defined as
	\[
	\hat{E} =\argmin \left\{
	-\frac{1}{|\hat{E}|}\sum_t \log \max_{i \not \in \hat{E}^{t-1}} \Pr(i; \hat{E}^{t-1}) \right\},
	\]
	where $\hat{E}^t$ is the predicted evidence until step $t$.
	QFE uses the beam search algorithm 
	to search $\hat{E}$.

	\section{Experiments on RC}
	\label{sec:RC}
	\subsection{HotpotQA Dataset}	
	In HotpotQA, the query $Q$ is created by crowd workers, on the condition that answering $Q$ requires reasoning over two paragraphs in Wikipedia. The candidates of $A_T$ are `Yes', `No', and `Span'. The answer string $A_S$, if it exists, is a span in the two paragraphs. 
	The context $C$ is ten paragraphs, and its content has two settings. In the \textbf{distractor setting}, $C$ consists of the two gold paragraphs used to create $Q$ and eight paragraphs retrieved from Wikipedia by using TF-IDF with $Q$.
	Table \ref{hotpot} shows the statistics of the distractor setting.
	In the \textbf{fullwiki setting}, all ten paragraphs of $C$ are retrieved paragraphs. Hence, $C$ may not include two gold paragraphs, and in that case, $A_S$ and $E$ cannot be extracted. 
	Therefore, the oracle model does not achieve 100 \% accuracy.
	HotpotQA does not provide the training data for the fullwiki setting, and the training data in the fullwiki setting is the same as the distractor setting.
	
	\begin{table}[t]
	\begin{center}
		\scalebox{0.8}{
			\begin{tabular}{rcccc}\hline
				& \multicolumn{2}{c}{Context} & Query & Evidence\\
				& \# paragraphs & \# words & \# words & \# sentences \\ \hline
				Ave. & 10.0 & 1162.0 & 17.8 & 2.4 \\ 
				Max & 10 & 3079 & 59 & 8 \\ 
				Median & 10 & 1142 & 17 & 2 \\
				Min & 2 & 60 & 7 & 2 \\ \hline
		\end{tabular}}
	\end{center}
	\caption{Statistics of HotpotQA (the development set in the distractor setting).}
	\label{hotpot}
	\end{table}

	\subsection{Experimental Setup}
	
    \paragraph{Comparison models}
    	Our baseline model is the same as the baseline in \citet{hotpot} except as follows. Whereas we use equation (\ref{eq:qfe}), they use
	\[
	\Pr(i) = \mathrm{sigmoid}(w^\top x_i+b),
	\]
	where $w\in \mathbb{R}^{2d_c}, b \in \mathbb{R}$ are trainable parameters. The evidence loss $L_E$ is the sum of binary cross-entropy functions on whether each of the sentences is evidence or not. In the test phase, the sentences with probabilities higher than a threshold are selected. We set the threshold to 0.4 because it gave the highest F1 score on the development set.
	The remaining parts of the implementations of our and baseline models are the same. The details are in Appendix A.1.	
	
	We also compared \textbf{DFGN + BERT}~\cite{dfgn}, \textbf{Cognitive Graph}~\cite{cognotive_graph}, \textbf{GRN} and \textbf{BERT Plus}, which were unpublished at the submission time (4 March 2019).
	
	\paragraph{Evaluation metrics}
	We evaluated the prediction of $A_T$, $A_S$ and $E$ by using the official metrics in HotpotQA. Exact match (EM) and partial match (F1) were used to evaluate both the answer and the evidence. 
	For the answer evaluation, 
	the score was measured by the classification accuracy of $A_T$.
	Only when $A_T$ was `Span' was the score also measured by the word-level matching of $A_S$.
	For the evidence, the partial match was evaluated by the sentence ids, 
	so word-level partial matches were not considered.
	For metrics on both the answer and the evidence,
	we used Joint EM and Joint F1 \cite{hotpot}.
	
	\subsection{Results}
	
	\paragraph{Does our model achieve state-of-the-art performance?}
	\begin{table}[t]
		\begin{center}
			\scalebox{0.75}{
				\begin{tabular}{r|c|c|c|c|c|c}\hline
					& \multicolumn{2}{c|}{Answer}
					& \multicolumn{2}{|c|}{Evidence}
					& \multicolumn{2}{|c}{Joint}			
					\\ \hline
					& EM & F1 & EM & F1 & EM & F1 \\ \hline
					Baseline & 45.6 & 59.0 & 20.3 & 64.5 & 10.8 & 40.2 \\ 
					BERT Plus & \textbf{56.0} & \textbf{69.9} & 42.3 & 80.6 & 26.9 & 58.1 \\
					DFGN + BERT & 55.2 & 68.5 & 49.9 & 81.1 & 31.9 & 58.2 \\
					GRN & 52.9 & 66.7 & 52.4 & 84.1 & 31.8 & 58.5 \\\hline
					QFE & 53.9 & 68.1 & \textbf{57.8}
					& \textbf{84.5} & \textbf{34.6} & \textbf{59.6} \\\hline
			\end{tabular}}
			\caption{
			Performance of the models on the HotpotQA distractor setting leaderboard\footnotemark[1] (4 March 2019). The models except for the baseline were unpublished at the time of submission of this paper. Our model was submitted on 21 November 2018, three months before the other submissions. }
			\label{tab:hotpot_test_dist}
			\end{center}
        \end{table}
        \begin{table}
        \begin{center}
			\scalebox{0.75}{
				\begin{tabular}{r|c|c|c|c|c|c}\hline
					& \multicolumn{2}{c|}{Answer}
					& \multicolumn{2}{|c|}{Evidence}
					& \multicolumn{2}{|c}{Joint}			
					\\ \hline
					& EM & F1 & EM & F1 & EM & F1 \\ \hline
					Baseline & 24.0 & 32.9 & 3.86 & 37.7 & 1.85 & 16.2 \\ 
					GRN & 27.3 & 36.5 & 12.2 & 48.8 & 7.40 & 23.6 \\
					Cognitive Graph & \textbf{37.1} & \textbf{48.9} & \textbf{22.8} 
					& \textbf{57.8} & \textbf{12.4} & \textbf{34.9} \\ \hline
					QFE & 28.7 & 38.1 & 14.2 & 44.4 & 8.69 & 23.1 \\ \hline
			\end{tabular}}
			\caption{Performance of the models on the HotpotQA fullwiki setting leaderboard\footnotemark[1]  (4 March 2019). The models except for the baseline were unpublished at the time of submission of this paper. Our model was submitted on 25 November 2018, three months before the other submissions.}
			\label{tab:hotpot_test_full}
		\end{center}
	\end{table}
    
    Table \ref{tab:hotpot_test_dist} shows that, in the distractor setting, 
    QFE performed the best in terms of the evidence extraction score among all models compared.
    It also achieved comparable performance in terms of the answer selection score and therefore achieved state-of-the-art performance on the joint EM and F1 metrics, which are the main metric on the dataset.
     QFE outperformed the 
     baseline model in all metrics.
    Although our model does not use any pre-trained language model such as BERT \cite{bert} for encoding, it outperformed the methods that used BERT such as DFGN + BERT and BERT Plus.
    In particular, the improvement in the evidence EM score was +37.5 points against the baseline and +5.4 points against GRN.
    
    In the fullwiki setting, Table \ref{tab:hotpot_test_full} shows that QFE 
    outperformed the baseline
    in all metrics.
    Compared with the unpublished model at the submission time, Cognitive Graph \cite{cognotive_graph} outperformed our model. 
    There is a dataset shift problem~\cite{dataset_shift} in HotpotQA, 
    where the distribution of the number of gold evidence sentences and the answerability
    differs between training (i.e., the distractor setting) and test (i.e., the fullwiki setting) phases. In the fullwiki setting, the questions may have less than two gold evidence sentences or be even unanswerable. %
    Our current QA and QFE models do not consider solving 
    the dataset shift problem; 
    our future work will deal with
    it.
    \footnotetext[1]{https://hotpotqa.github.io/}
	
	\paragraph{Does QFE contribute to the performance?}
	\begin{table}[t]
	\begin{center}
		\scalebox{0.75}{
			\begin{tabular}{r|c|c|c|c|c|c}\hline
				& \multicolumn{2}{c|}{Answer}
				& \multicolumn{2}{|c|}{Evidence}
				& \multicolumn{2}{|c}{Joint}			
				\\ \hline
				& EM & F1 & EM & F1 & EM & F1 \\ \hline
				\citet{hotpot} & 44.4 & 58.3 & 22.0 & 66.7 & 11.6 & 40.9 \\ 
				our implementation\footnotemark[2] & 52.7 & 67.3 & 38.0 & 78.4 & 21.9 & 54.9 \\
				+ top 2 extraction & 52.7 & 67.3 & 48.0 & 77.8 & 27.6 & 54.4 \\ \hline
				QFE &  \textbf{53.7} & \textbf{68.7} 
				& \textbf{58.8} & \textbf{84.7} & \textbf{35.4} 
				& \textbf{60.6} \\ 
				without glimpse & 53.1 & 67.9 & 58.4 & 84.3 & 34.8 & 59.6 \\
				pipeline model & 46.9 & 63.6 & -- & -- & -- & -- \\\hline 
		\end{tabular}}
		\caption{Performance of our models and the baseline models on the development set in the distractor setting.
		}
		\label{tab:hotpot_dev_dist} 
	\end{center}
	\end{table}
    \footnotetext[2]{The differences in score among the original and our implementations of \citet{hotpot} are due to the hyper parameters. The main change is increasing $d_c$ from 50 to 150.}
    
	Table \ref{tab:hotpot_dev_dist} shows the results of the ablation study. 
	
	QFE performed the best among the models compared. Although the difference between our overall model and the baseline is the evidence extraction model, the answer scores also improved. 
	QFE also outperformed the model that used only RNN extraction without glimpse.
	
	QFE defines the terminal condition as reaching the EOE sentence, which we call adaptive termination.
	We confirmed that the adaptive termination of QFE contributed to its performance. We compared QFE with a baseline that extracts the two sentences with the highest scores, since the most frequent number of evidence sentences is two. QFE outperformed this baseline. 
	
	Our model uses the results of evidence extraction as a guide for selecting the answer, but it is not a pipeline model of evidence extraction and answer selection. 
	Therefore, we evaluated a pipeline model that selects the answer string $A_S$ only from the extracted evidence sentences, where the outputs of the answer layer corresponding to non-evidence sentences are masked with the prediction of the evidence extraction.
	Although almost all answer strings in the dataset are in the gold evidence sentences,
	the model performed poorly.
    We consider that the evidence extraction helps QA model to learn, but its performance is not enough to improve the performance of the answer layer with the pipeline model.

	\paragraph{What are the characteristics of our evidence extraction?}
	\begin{table}[t]
    \begin{center}
    	\scalebox{0.9}{
			\begin{tabular}{rccc}\hline
                & Precision & Recall & Correlation \\ \hline
				baseline & 79.0 & 82.4 & 0.259 \\ 
				QFE  & \textbf{88.4} & \textbf{83.2} & \textbf{0.375} \\ \hline
		\end{tabular}}
		\caption{Performance of our model and the baseline in evidence extraction on the development set in the distractor setting. The correlation is the Kendall tau correlation of the number of predicted evidence sentences and that of gold evidence.} 
		\label{tab:hotpot_evi}
	\end{center}
	\end{table}
	\begin{figure}[t]
	\begin{center}
		\includegraphics[width =75mm]{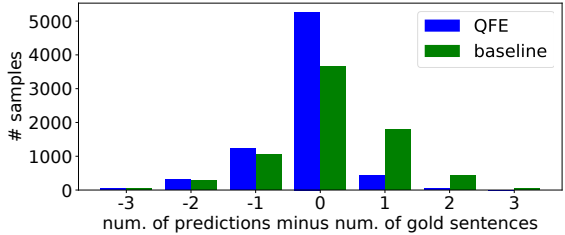}
		\caption{Number of predicted evidence sentences minus the number of gold evidence sentences.}
		\label{fig:hist_dist}
	\end{center}
	\end{figure}

	Table \ref{tab:hotpot_evi} shows the evidence extraction performance in the distractor setting. Our model improves both precision and recall, and the improvement in precision is larger.
	
	Figure \ref{fig:hist_dist} reveals the reason for the high EM and precision scores; QFE rarely extracts too much evidence. That is, it predicts the number of evidence sentences  more accurately than the baseline. Table \ref{tab:hotpot_evi} also shows the correlation of our model about the number of evidence sentences is higher than that of the baseline.
	
	We consider that the sequential extraction and the adaptive termination help to prevent over-extraction. In contrast, the baseline evaluates each sentence independently, so the baseline often extracts too much evidence.

	\paragraph{What questions in HotpotQA are difficult for QFE?}
		\begin{table}[t]
		\begin{center}
    		\scalebox{0.7}{
				\begin{tabular}{r|c|c|c|c|c|c|c|c}\hline
				    \multicolumn{2}{c}{}
					& \multicolumn{2}{|c}{Answer}
					& \multicolumn{5}{|c}{Evidence}
					\\ \hline
					\# Evi & \# sample & EM & F1 & Num & EM & P & R & F1 \\ \hline
				    all & 100 & 53.7 & 68.7 & 2.22 & 58.8 & 88.4 & 83.2 & 84.7 \\ \hline
					2 & 67.4 & 54.8 & 69.6 & 2.09 & 76.9 & 88.4 & 91.1 & 89.4 \\ 
					3 & 24.0 & 52.5 & 68.4 & 2.43 & 26.0 & 89.3 & 71.8 & 78.7 \\ 
					4 & 7.25 & 52.5  & 66.9 & 2.61 & 14.0 & 90.7 & 59.4 & 70.4 \\ 
					5 & 1.08 & 42.5 & 57.0 & 2.65 & 2.50 & 92.1 & 49.5 & 63.1 \\ \hline
					\end{tabular}}
			\caption{Performance of our model in terms of the number of gold evidence sentences on the development set in the distractor setting. \# sample, Num, P and R mean the proportion in the dataset, number of predicted evidence sentences, precision, and recall, respectively.}
			\label{tab:hotpot_length}
		\end{center}
	\end{table}
	\begin{table}[t]
	    \begin{center}
			\scalebox{0.85}{
				\begin{tabular}{r|c|c|c|c|c|c}\hline
					& \multicolumn{2}{c|}{Answer}
					& \multicolumn{2}{|c|}{Evidence}
					& \multicolumn{2}{|c}{Joint}			
					\\ \hline
					& EM & F1 & EM & F1 & EM & F1 \\ \hline
				    all & 53.7 & 68.7 & 58.8 & 84.7 & 35.4 & 60.6 \\ \hline
					comparison & 54.1 & 60.7 & 71.2 & 88.8 & 42.0 & 55.6 \\ 
					bridge & 53.6 & 70.7 & 55.7 & 83.7 &  33.8& 61.8 \\\hline
			\end{tabular}}
			\caption{Performance of our model for each reasoning type
			on the development set in the distractor setting.} 
			\label{tab:hotpot_type}
		\end{center}
	\end{table}
				\begin{table*}[t]
		\begin{center}
			\scalebox{0.85}{
			\begin{tabular}{ccll}\hline
					\multicolumn{4}{l}{$Q$: Which band has more members, Kitchens of Distinction or Royal Blood?
					\quad $A_T=\hat{A}_T$: Kitchens of Distinction } \\ \hline
					gold & predicted & probability[\%] & text \\ \hline
					\checkmark & 1 & \textbf{96.9} & 
					\parbox{30em}{\strut{}Kitchens of Distinction ... are an English three-person alternative rock band ...
			        \strut} \\ 
					\checkmark & 2 & $0.2 \to \textbf{81.4}$ & 
					\parbox{30em}{\strut{}Royal Blood are an English rock duo formed in Brighton in 2013.
			        \strut} \\ 
			         & 3 & $0.0 \to ~~0.0 \to \textbf{52.3}$ & 
					\parbox{30em}{\strut{}EOE sentence \strut} \\ 
					 & --- & $2.9 \to 16.8 \to 31.9$  & 
					\parbox{30em}{\strut{}In September 2012, ... members ... as Kitchens of Distinction.
			        \strut} \\
			         & --- & $0.0 \to ~~0.0 \to ~~0.0$ & 
			        \parbox{30em}{\strut{}Royal Blood is the eponymous debut studio album by British rock duo Royal Blood.
			        \strut} \\ \hline
			\end{tabular}}
		\end{center}
		\caption{Outputs of QFE. The sentences are extracted in the order shown in the predicted column. The extraction scores of the sentences at each step are in the probability column.}
		\label{tab:error}
	\end{table*}
	
	We analyzed the difficulty of the questions for QFE from the perspective of the number of evidence sentences and reasoning type; the results are in Table \ref{tab:hotpot_length} and Table \ref{tab:hotpot_type}.
	
	First, we classified the questions by the number of gold evidence sentences. Table \ref{tab:hotpot_length} shows the model performance for each number. The answer scores were low for the questions answered with five evidence sentences, which indicated that questions requiring much evidence are difficult. However, the five-evidence questions amount to only 80 samples, so this observation needs to be confirmed with more analysis.
	QFE performed well when the number of gold evidence sentences was two.
	Even though QFE was relatively conservative when extracting many evidence sentences, it was able to extract more than two sentences adaptively.
	
    Second, we should mention the reasoning types in Table \ref{tab:hotpot_type}. HotpotQA has two reasoning types: entity bridge and entity comparison. Entity bridge means that the question mentioned one entity and the article of this entity has another entity required for the answer. Entity comparison means that the question compares two entities. 
    
    Table \ref{tab:hotpot_type} shows that QFE works on each reasoning type. 
	We consider that the difference between the results is due to the characteristics of the dataset.
	The answer F1 was relatively low in the comparison questions, 
	because all yes/no questions belong to the comparison question
	and partial matches do not happen in yes/no questions.
	The evidence EM was relatively high in the comparison questions.
	One of the reason is that 77.1 \% of the comparison questions have just two evidence sentences.
	This proportion is larger than that in the bridge questions, 64.9\%.
	From another perspective, the comparison question sentence itself will contain the clues (i.e., two entities) required to gather all evidence sentences, while the bridge question sentence itself will provide only a part of the clues and require multi-hop reasoning, i.e., finding an evidence sentence from another evidence sentence.
	Therefore, the evidence extraction of the bridge questions is more difficult than that of the comparison questions.

	\paragraph{Qualitative Analysis.}
    Table \ref{tab:error} shows an example of the behavior of QFE. In it, the system must compare the number of members of Kitchens of Distinction and with those of Royal Blood. The system extracted the two sentences describing the number of members.
	Then, the system extracted the EOE sentence.
	
	We should note two sentences that were not extracted. The first sentence includes `members' and `Kitchens of Distinction', which are included in the query. However, this sentence does not mention the number of the members of Kitchens of Distinction. The second sentence also shows that Royal Blood is a duo.
	However, our model preferred Royal Blood (band name) to Royal Blood (album name) as the subject of the sentence.
	
	Other examples are shown in Appendix A.2.
	
	\section{Experiments on RTE}
	\subsection{FEVER Dataset}	
	In FEVER, the query $Q$ is created by crowd workers. Annotators are given a randomly sampled sentence and a corresponding dictionary. The given sentence is from Wikipedia. The key-value of the corresponding dictionary consists of an entity and a description of the entity. Entities are those that have a hyperlink from the given sentence. The description is the first sentence of the entity's Wikipedia page. Only using the information in the sentence and the dictionary, annotators create a claim as $Q$. The candidates of $A_T$ are `Supports', `Refutes' and `Not Enough Info (NEI)'. The proportion of samples with more than one evidence sentence is 27.3\% in the samples whose label is not `NEI'.
	The context $C$ is the Wikipedia database shared among all samples.
	Table \ref{fever} shows the statistics.
	
	\begin{table}[t]
		\begin{center}
			\scalebox{0.9}{
				\begin{tabular}{rccc}\hline
					& Context & Query & Evidence\\
					& \# pages & \# words & \# sentences \\ \hline
					Ave. & 5416537 & 9.60 & 1.13 \\ 
					Max & --- & 39 & 52 \\ 
					Median & --- & 9 & 1 \\
					Min & --- & 3 & 0 \\ \hline
			\end{tabular}}
		\end{center}
		\caption{Statistics of FEVER (the development set).}
		\label{fever}
	\end{table}

	\subsection{Experimental Setup}
	
	Because $C$ is large, we used the NSMN document retriever \cite{fever1} and gave only the top-five paragraphs to our model. Similar to NSMN, in order to capture the semantic and numeric relationships, we used 30-dimensional WordNet features and five-dimensional number embeddings. The WordNet features are binaries reflecting the existence of hypernymy/antonymy words in the input. The number embedding is a real-valued embedding assigned to any unique number.

	Because the number of samples in the training data is biased on the answer type $A_T$, randomly selected samples were copied in order to equalize the numbers. 
	Our model used ensemble learning of 11 randomly initialized models. For the evidence extraction, we used the union of the predicted evidences of each model. 
	If the model predicts $A_T$ as `Supports' or `Refutes', the model extracts at least one sentence.
	Details of the implementation are in Appendix A.1.

	We evaluated the prediction of $A_T$ and the evidence $E$ by using the official metrics in FEVER. $A_T$ was evaluated in terms of the label accuracy. $E$ was evaluated in terms of precision, recall and F1, which were measured by sentence id. The FEVER score was used as a metric accounting for both $A_T$ and $E$. The FEVER score of a sample is 1 if the predicted evidence includes all gold evidence and the answer is correct. That is, the FEVER score emphasizes the recall of extracting evidence sentences over the precision.

	\subsection{Results}
	\begin{table}[t]
	\begin{center}
		\scalebox{0.85}{
			\begin{tabular}{rccc}\hline
				& Evidence
				& Answer
				& FEVER			
				\\
				& F1 & Acc. &  \\ \hline
				\citet{fever1} & 53.0 & 68.2 & 64.2 \\
				\citet{fever2} & 35.0 & 67.6 & 62.5 \\ \hline
			    who & 37.4 & \textbf{72.1} & \textbf{66.6} \\
			    Kudo & 36.8 & 70.6 & 65.7 \\
			    avonamila & 60.3 & 71.4 & 65.3 \\ \hline
			    hz66pasa & 71.4 & 33.3 & 22.0 \\
				aschern & 70.4 & 69.3 & 60.9 \\ \hline
				QFE & \textbf{77.7} & 69.3 & 61.8 \\\hline
			\end{tabular}}
		\caption{Performance of the models on the FEVER leaderboard\footnotemark[3] (4 March 2019).
		The top two rows are the models submitted during the FEVER Shared Task that have higher FEVER scores than ours. The middle three rows are the top-three FEVER models submitted after the Shared Task. 
		The rows next to the bottom and the bottom row (ours) show the top-three F1 models submitted after the Shared Task. None of the models submitted after the Shared Task has paper information.
		}
		\label{tab:fever_test}
	\end{center}
	\end{table}
	\footnotetext[3]{https://competitions.codalab.org/competitions/18814}
	\begin{table}[t]
	\begin{center}
			\scalebox{0.85}{
			\begin{tabular}{rccc}\hline
					& Precision & Recall & F1  \\ \hline
					\citet{fever1} & 42.3 & 70.9 & 53.0 \\
					\citet{fever2} & 22.2 & 82.8 & 35.0 \\
					\citet{fever3} & 23.6 & \textbf{85.2} & 37.0 \\
					\citet{fever4} & \textbf{92.2} & 50.0 & 64.9 \\ \hline
					QFE ensemble (test) & 79.1 & 76.3 & \textbf{77.7} \\ \hline \hline
					QFE single (dev) & 90.8 & 64.9 & 76.6 \\
					QFE ensemble (dev) & 83.9 & 78.1 & 81.0 \\ \hline
			\end{tabular}}
		\caption{Performance of evidence extraction. The top five rows are evaluated on the test set. The comparison of our models is on the development set. The models submitted after the Shared Task have no information about precision or recall.
		}
		\label{tab:fever_evi}
	\end{center}
	\end{table}
	
	\paragraph{Does our multi-task learning approach achieve state-of-the-art performance?}
	Table \ref{tab:fever_test} shows QFE achieved state-of-the-art performance in terms of the evidence F1 and comparable performance in terms of label accuracy to the competitive models.
	The FEVER score of our model is lower than those of other models, because the FEVER score emphasizes recall. However, the importance of the precision and the recall depends on the utilization. QFE is suited to situations where concise output is preferred.

    \paragraph{What are the characteristics of our evidence extraction?}
    Table \ref{tab:fever_evi} shows our model achieved high performance on all metrics of evidence extraction. On the test set, it ranked in 2nd place in precision, 3rd place in recall, and 1st place in F1. As for the results on the development set, QFE extracted with higher precision than recall. This tendency was the same as in the RC evaluation. The single model has a larger difference between precision and recall. The ensemble model improves recall and F1.
    
    Examples are shown in Appendix A.2.

	\section{Related Work}
	\subsection{Reading Comprehension}
	RC is performed by matching the context and the query \cite{bidaf}. Many RC datasets referring to multiple texts have been published, such as MS MARCO \cite{marco} and TriviaQA \cite{triviaqa}. For such datasets, the document retrieval model is combined with the context-query matching model \cite{drqa, R3, evidence, CIKM}.
	
	Some techniques have been proposed for understanding multiple texts.
	\citet{simple} used simple methods, such as connecting texts. \citet{coarse1, coarse2} proposed a combination of coarse reading and fine reading.
	However, \citet{RCanal} indicated that
	most questions in RC require reasoning from just one sentence including the answer.
	The proportion of such questions is more than 63.2 \% in TriviaQA and 86.2 \% in MS MARCO.
	
	This observation is one of the motivations behind multi-hop QA.
	HotpotQA \cite{hotpot} is a task including supervised evidence extraction. QAngaroo \cite{qangaroo} is a task created by using Wikipedia entity links. The difference between QAngaroo and our focus is two-fold: (1) QAngaroo does not have supervised evidence and (2) the questions in QAngaroo are inherently limited because the dataset is constructed using a knowledge base.
	MultiRC \cite{multirc} is also an explainable multi-hop QA dataset that provides gold evidence sentences.
	However, 
	it is difficult to compare the performance of the evidence extraction with other studies because its evaluation script and leaderboard do not report the evidence extraction score.
	
	Because annotation of the evidence sentence is costly, 
	unsupervised learning of the evidence extraction is another important issue.
	\citet{unsupervised_evidence} tackled unsupervised learning for explainable multi-hop QA,
	but their model is restricted to the multiple-choice setting.
	
	\subsection{Recognizing Textual Entailment}
	RTE \cite{snli, mnli} is performed by sentence matching \cite{Rock, esim}.
	
	FEVER \cite{fever} has the aim of verification and fact checking for RTE on a large database. FEVER requires three sub tasks: document retrieval, evidence extraction, and answer prediction. 
	In the previous work, the sub tasks are performed using pipelined models \cite{fever1, fever2}. 
	In contrast, our approach performs evidence extraction and answer prediction simultaneously by regarding FEVER as an explainable multi-hop QA task.

	\subsection{Summarization}
	A typical approach to sentence-level extractive summarization has an encoder-decoder architecture \cite{seq2seq,summarunner,refresh}. 
	Sentence-level extractive summarization is also used for content selection in abstractive summarization \citep{fast}.
	The model extracts sentences in order of importance and edits them.  We have extended this model so that it can be used for evidence extraction because we consider that the evidence must be extracted in order of importance rather than the original order, which the conventional models use.

	\section{Conclusion}
	We consider that the main contributions of our study are 
	(1) the QFE model that is based on a summarization model for the explainable multi-hop QA, 
	(2) the dependency among the evidence and the coverage of the question due to the usage of the summarization model, 
	and (3) the state-of-the-art performance in evidence extraction in both RC and RTE tasks.
	
	Regarding RC, we confirmed that the architecture with QFE, which is a simple replacement of the baseline, achieved state-of-the-art performance in the task setting.
	The ablation study showed that the replacement of the evidence extraction model with QFE improves performance.
	Our adaptive termination contributes to the exact matching and the precision score of the evidence extraction.
	The difficulty of the questions for QFE depends on the number of the required evidence sentences.
	This study is the first to base its experimental discussion on HotpotQA.
	
	Regarding RTE, we confirmed that, compared with competing models, the architecture with QFE has a higher evidence extraction score and comparable label prediction score.
	This study is the first to show a joint approach for RC and FEVER.
	
	\bibliography{theme}

\begin{thebibliography}{40}
\expandafter\ifx\csname natexlab\endcsname\relax\def\natexlab#1{#1}\fi

\bibitem[{Bowman et~al.(2015)Bowman, Angeli, Potts, and Manning}]{snli}
Samuel~R. Bowman, Gabor Angeli, Christopher Potts, and Christopher~D. Manning.
  2015.
\newblock \href {https://www.aclweb.org/anthology/D15-1075} {A large annotated
  corpus for learning natural language inference}.
\newblock In \emph{EMNLP}, pages 632--642.

\bibitem[{Chen et~al.(2017{\natexlab{a}})Chen, Fisch, Weston, and
  Bordes}]{drqa}
Danqi Chen, Adam Fisch, Jason Weston, and Antoine Bordes. 2017{\natexlab{a}}.
\newblock \href {https://www.aclweb.org/anthology/P17-1171} {{Reading Wikipedia
  to Answer Open-Domain Questions}}.
\newblock In \emph{ACL}, pages 1870--1879.

\bibitem[{Chen et~al.(2017{\natexlab{b}})Chen, Zhu, Ling, Wei, Jiang, and
  Inkpen}]{esim}
Qian Chen, Xiaodan Zhu, Zhenhua Ling, Si~Wei, Hui Jiang, and Diana Inkpen.
  2017{\natexlab{b}}.
\newblock \href {https://www.aclweb.org/anthology/P17-1152} {Enhanced {LSTM}
  for natural language inference}.
\newblock In \emph{ACL}, pages 1657--1668.

\bibitem[{Chen and Bansal(2018)}]{fast}
Yen-Chun Chen and Mohit Bansal. 2018.
\newblock \href {https://www.aclweb.org/anthology/P18-1063} {Fast abstractive
  summarization with reinforce-selected sentence rewriting}.
\newblock In \emph{ACL}, pages 675--686.

\bibitem[{Cheng and Lapata(2016)}]{seq2seq}
Jianpeng Cheng and Mirella Lapata. 2016.
\newblock \href {https://www.aclweb.org/anthology/P16-1046} {Neural
  summarization by extracting sentences and words}.
\newblock In \emph{ACL}, pages 484--494.

\bibitem[{Cho et~al.(2014)Cho, Merrienboer, Gulcehre, Bougares, Schwenk, and
  Bengio}]{gru}
Kyunghyun Cho, Bart Merrienboer, Caglar Gulcehre, Fethi Bougares, Holger
  Schwenk, and Yoshua Bengio. 2014.
\newblock \href {https://www.aclweb.org/anthology/D14-1179} {Learning phrase
  representations using rnn encoder-decoder for statistical machine
  translation}.
\newblock In \emph{EMNLP}, pages 1724--1734.

\bibitem[{Choi et~al.(2017)Choi, Hewlett, Uszkoreit, Polosukhin, Lacoste, and
  Berant}]{coarse1}
Eunsol Choi, Daniel Hewlett, Jakob Uszkoreit, Illia Polosukhin, Alexandre
  Lacoste, and Jonathan Berant. 2017.
\newblock \href {https://www.aclweb.org/anthology/P17-1020} {Coarse-to-fine
  question answering for long documents}.
\newblock In \emph{ACL}, pages 209--220.

\bibitem[{Clark and Gardner(2018)}]{simple}
Christopher Clark and Matt Gardner. 2018.
\newblock \href {https://www.aclweb.org/anthology/P18-1078} {Simple and
  effective multi-paragraph reading comprehension}.
\newblock In \emph{ACL}, pages 845--855.

\bibitem[{Devlin et~al.(2019)Devlin, Chang, Lee, and Toutanova}]{bert}
Jacob Devlin, Ming-Wei Chang, Kenton Lee, and Kristina Toutanova. 2019.
\newblock {BERT}: Pre-training of deep bidirectional transformers for language
  understanding.
\newblock \emph{NAACL-HLT}.
\newblock To appear.

\bibitem[{Ding et~al.(2019)Ding, Zhou, Chen, Yang, and Tang}]{cognotive_graph}
Ming Ding, Chang Zhou, Qibin Chen, Hongxia Yang, and Jie Tang. 2019.
\newblock Cognitive graph for multi-hop reading comprehension at scale.
\newblock In \emph{ACL}.
\newblock To appear.

\bibitem[{Hanselowski et~al.(2018)Hanselowski, Zhang, Li, Sorokin, Schiller,
  Schulz, and Gurevych}]{fever3}
Andreas Hanselowski, Hao Zhang, Zile Li, Daniil Sorokin, Benjamin Schiller,
  Claudia Schulz, and Iryna Gurevych. 2018.
\newblock \href {https://www.aclweb.org/anthology/W18-5516} {Ukp-athene:
  Multi-sentence textual entailment for claim verification}.
\newblock In \emph{FEVER@EMNLP}, pages 103--108.

\bibitem[{Joshi et~al.(2017)Joshi, Choi, Weld, and Zettlemoyer}]{triviaqa}
Mandar Joshi, Eunsol Choi, Daniel~S Weld, and Luke Zettlemoyer. 2017.
\newblock \href {https://www.aclweb.org/anthology/P17-1147} {{TriviaQA}: A
  large scale distantly supervised challenge dataset for reading
  comprehension}.
\newblock In \emph{ACL}, pages 1601--1611.

\bibitem[{Khashabi et~al.(2018)Khashabi, Chaturvedi, Roth, Upadhyay, and
  Roth}]{multirc}
Daniel Khashabi, Snigdha Chaturvedi, Michael Roth, Shyam Upadhyay, and Dan
  Roth. 2018.
\newblock \href {https://www.aclweb.org/anthology/N18-1023} {Looking beyond the
  surface: A challenge set for reading comprehension over multiple sentences}.
\newblock In \emph{NAACL-HLT}, pages 252--262.

\bibitem[{Kim(2014)}]{char}
Yoon Kim. 2014.
\newblock \href {https://www.aclweb.org/anthology/D14-1181} {Convolutional
  neural networks for sentence classification}.
\newblock In \emph{EMNLP}, pages 1746--1751.

\bibitem[{Kingma and Ba(2014)}]{adam}
Diederik~P. Kingma and Jimmy Ba. 2014.
\newblock Adam: {A} method for stochastic optimization.
\newblock \emph{arXiv preprint arXiv:1412.6980}.

\bibitem[{Malon(2018)}]{fever4}
Christopher Malon. 2018.
\newblock \href {https://www.aclweb.org/anthology/W18-5517} {Team papelo:
  Transformer networks at fever}.
\newblock In \emph{FEVER@EMNLP}, pages 109--113.

\bibitem[{Nallapati et~al.(2017)Nallapati, Zhai, and Zhou}]{summarunner}
Ramesh Nallapati, Feifei Zhai, and Bowen Zhou. 2017.
\newblock Summarunner: A recurrent neural network based sequence model for
  extractive summarization of documents.
\newblock In \emph{AAAI}, pages 3075--3081.

\bibitem[{Narayan et~al.(2018)Narayan, Cohen, and Lapata}]{refresh}
Shashi Narayan, Shay~B. Cohen, and Mirella Lapata. 2018.
\newblock \href {https://www.aclweb.org/anthology/N18-1158} {Ranking sentences
  for extractive summarization with reinforcement learning}.
\newblock In \emph{NAACL-HLT}, pages 1747--1759.

\bibitem[{Nguyen et~al.(2016)Nguyen, Rosenberg, Song, Gao, Tiwary, Majumder,
  and Deng}]{marco}
Tri Nguyen, Mir Rosenberg, Xia Song, Jianfeng Gao, Saurabh Tiwary, Rangan
  Majumder, and Li~Deng. 2016.
\newblock {MS MARCO}: A human generated machine reading comprehension dataset.
\newblock In \emph{CoCo@NIPS}.

\bibitem[{Nie et~al.(2019)Nie, Chen, and Bansal}]{fever1}
Yixin Nie, Haonan Chen, and Mohit Bansal. 2019.
\newblock Combining fact extraction and verification with neural semantic
  matching networks.
\newblock In \emph{AAAI}.
\newblock To appear.

\bibitem[{Nishida et~al.(2018)Nishida, Saito, Otsuka, Asano, and Tomita}]{CIKM}
Kyosuke Nishida, Itsumi Saito, Atsushi Otsuka, Hisako Asano, and Junji Tomita.
  2018.
\newblock {Retrieve-and-Read: Multi-task Learning of Information Retrieval and
  Reading Comprehension}.
\newblock In \emph{CIKM}, pages 647--656.

\bibitem[{Pennington et~al.(2014)Pennington, Socher, and Manning}]{glove}
Jeffrey Pennington, Richard Socher, and Christopher~D. Manning. 2014.
\newblock \href {https://www.aclweb.org/anthology/D14-1162} {Glo{V}e: Global
  vectors for word representation}.
\newblock In \emph{EMNLP}, pages 1532--1543.

\bibitem[{Quionero-Candela et~al.(2009)Quionero-Candela, Sugiyama,
  Schwaighofer, and D.~Lawrence}]{dataset_shift}
Joaquin Quionero-Candela, Masashi Sugiyama, Anton Schwaighofer, and Neil
  D.~Lawrence. 2009.
\newblock \emph{Dataset Shift in Machine Learning}.
\newblock The MIT Press.

\bibitem[{Rajpurkar et~al.(2016)Rajpurkar, Zhang, Lopyrev, and Liang}]{squad}
Pranav Rajpurkar, Jian Zhang, Konstantin Lopyrev, and Percy Liang. 2016.
\newblock \href {https://www.aclweb.org/anthology/D16-1264} {{SQuAD}: 100,000+
  questions for machine comprehension of text}.
\newblock In \emph{EMNLP}, pages 2383--2392.

\bibitem[{Rockt{\"a}schel et~al.(2016)Rockt{\"a}schel, Grefenstette, Hermann,
  Ko{\v{c}}isk{\`y}, and Blunsom}]{Rock}
Tim Rockt{\"a}schel, Edward Grefenstette, Karl~Moritz Hermann, Tom{\'a}{\v{s}}
  Ko{\v{c}}isk{\`y}, and Phil Blunsom. 2016.
\newblock Reasoning about entailment with neural attention.
\newblock In \emph{ICLR}.

\bibitem[{See et~al.(2017)See, Liu, and Manning}]{pointer}
Abigail See, Peter~J Liu, and Christopher~D Manning. 2017.
\newblock \href {https://www.aclweb.org/anthology/P17-1099} {Get to the point:
  Summarization with pointer-generator networks}.
\newblock In \emph{ACL}, pages 1073--1083.

\bibitem[{Seo et~al.(2017)Seo, Kembhavi, Farhadi, and Hajishirzi}]{bidaf}
Minjoon Seo, Aniruddha Kembhavi, Ali Farhadi, and Hannaneh Hajishirzi. 2017.
\newblock Bidirectional attention flow for machine comprehension.
\newblock In \emph{ICLR}.

\bibitem[{Sugawara et~al.(2018)Sugawara, Inui, Sekine, and Aizawa}]{RCanal}
Saku Sugawara, Kentaro Inui, Satoshi Sekine, and Akiko Aizawa. 2018.
\newblock \href {https://www.aclweb.org/anthology/D18-1453} {What makes reading
  comprehension questions easier?}
\newblock In \emph{EMNLP}, pages 4208--4219.

\bibitem[{Thorne et~al.(2018)Thorne, Vlachos, Cocarascu, Christodoulopoulos,
  and Mittal}]{fever}
James Thorne, Andreas Vlachos, Oana Cocarascu, Christos Christodoulopoulos, and
  Arpit Mittal. 2018.
\newblock \href {https://www.aclweb.org/anthology/W18-5501} {The fact
  extraction and verification ({FEVER}) shared task}.
\newblock In \emph{FEVER@EMNLP}, pages 1--9.

\bibitem[{Vinyals et~al.(2016)Vinyals, Bengio, and Kudlur}]{glimpse}
Oriol Vinyals, Samy Bengio, and Manjunath Kudlur. 2016.
\newblock Order matters: Sequence to sequence for sets.
\newblock In \emph{ICLR}.

\bibitem[{Wang et~al.(2019)Wang, Yu, Sun, Chen, Yu, Roth, and
  McAllester}]{unsupervised_evidence}
Hai Wang, Dian Yu, Kai Sun, Jianshu Chen, Dong Yu, Dan Roth, and David
  McAllester. 2019.
\newblock Evidence sentence extraction for machine reading comprehension.
\newblock \emph{arXiv preprint arXiv:1902.08852}.

\bibitem[{Wang et~al.(2018{\natexlab{a}})Wang, Yu, Guo, Wang, Klinger, Zhang,
  Chang, Tesauro, Zhou, and Jiang}]{R3}
Shuohang Wang, Mo~Yu, Xiaoxiao Guo, Zhiguo Wang, Tim Klinger, Wei Zhang, Shiyu
  Chang, Gerald Tesauro, Bowen Zhou, and Jing Jiang. 2018{\natexlab{a}}.
\newblock R3: Reinforced reader-ranker for open-domain question answering.
\newblock In \emph{AAAI}, pages 5981--5988.

\bibitem[{Wang et~al.(2018{\natexlab{b}})Wang, Yu, Jiang, Zhang, Guo, Chang,
  Wang, Klinger, Tesauro, and Campbell}]{evidence}
Shuohang Wang, Mo~Yu, Jing Jiang, Wei Zhang, Xiaoxiao Guo, Shiyu Chang, Zhiguo
  Wang, Tim Klinger, Gerald Tesauro, and Murray Campbell. 2018{\natexlab{b}}.
\newblock Evidence aggregation for answer re-ranking in open-domain question
  answering.
\newblock In \emph{ICLR}.

\bibitem[{Wang et~al.(2017)Wang, Yang, Wei, Chang, and Zhou}]{self}
Wenhui Wang, Nan Yang, Furu Wei, Baobao Chang, and Ming Zhou. 2017.
\newblock \href {https://www.aclweb.org/anthology/P17-1018} {Gated
  self-matching networks for reading comprehension and question answering}.
\newblock In \emph{ACL}, pages 189--198.

\bibitem[{Welbl et~al.(2018)Welbl, Stenetorp, and Riedel}]{qangaroo}
Johannes Welbl, Pontus Stenetorp, and Sebastian Riedel. 2018.
\newblock \href {https://www.aclweb.org/anthology/Q18-1021} {Constructing
  datasets for multi-hop reading comprehension across documents}.
\newblock \emph{TACL}, 6:287--302.

\bibitem[{Williams et~al.(2018)Williams, Nangia, and Bowman}]{mnli}
Adina Williams, Nikita Nangia, and Samuel~R. Bowman. 2018.
\newblock \href {https://www.aclweb.org/anthology/N18-1101} {A broad-coverage
  challenge corpus for sentence understanding through inference}.
\newblock In \emph{NAACL-HLT}, pages 1112--1122.

\bibitem[{Xiao et~al.(2019)Xiao, Qu, Qiu, Zhou, Li, Zhang, and Yu}]{dfgn}
Yunxuan Xiao, Yanru Qu, Lin Qiu, Hao Zhou, Lei Li, Weinan Zhang, and Yong Yu.
  2019.
\newblock Dynamically fused graph network for multi-hop reasoning.
\newblock In \emph{ACL}.
\newblock To appear.

\bibitem[{Yang et~al.(2018)Yang, Qi, Zhang, Bengio, Cohen, Salakhutdinov, and
  Manning}]{hotpot}
Zhilin Yang, Peng Qi, Saizheng Zhang, Yoshua Bengio, William~W Cohen, Ruslan
  Salakhutdinov, and Christopher~D Manning. 2018.
\newblock \href {https://www.aclweb.org/anthology/D18-1259} {{HotpotQA}: A
  dataset for diverse, explainable multi-hop question answering}.
\newblock In \emph{EMNLP}, pages 2369--2380.

\bibitem[{Yoneda et~al.(2018)Yoneda, Mitchell, Welbl, Stenetorp, and
  Riedel}]{fever2}
Takuma Yoneda, Jeff Mitchell, Johannes Welbl, Pontus Stenetorp, and Sebastian
  Riedel. 2018.
\newblock \href {https://www.aclweb.org/anthology/W18-5515} {{UCL Machine
  Reading Group}: Four factor framework for fact finding {(HexaF)}}.
\newblock In \emph{FEVER@EMNLP}, pages 97--102.

\bibitem[{Zhong et~al.(2019)Zhong, Xiong, Keskar, and Socher}]{coarse2}
Victor Zhong, Caiming Xiong, Nitish~Shirish Keskar, and Richard Socher. 2019.
\newblock Coarse-grain fine-grain coattention network for multi-evidence
  question answering.
\newblock In \emph{ICLR}.

\end{thebibliography}
	\bibliographystyle{acl_natbib}
	
	\appendix
	\section{Supplemental Material}
	\subsection{Details of the Implementation \label{sec:implement}}
	We implemented our model in PyTorch and trained it on four Nvidia Tesla P100 GPUs. The RNN was a gated recurrent unit (GRU) \citep{gru}. The optimizer was Adam \cite{adam}. The word-based word embeddings were fixed GloVe 300-dimensional vectors \cite{glove}. The character-based word embeddings were obtained using trainable eight-dimensional character embeddings and a 100-dimensional CNN and max pooling. Table \ref{tab:hypara} shows other hyper parameters.
	
	In FEVER, if the model predicts $A_T$ as `Supports' or `Refutes', the model extracts at least one sentence by removing the EOE sentence from the candidates to be extracted at $t =1$. 
	
    \subsection{Samples of QFE Outputs \label{sec:sample}}
    The section describes some examples of QFE outputs. Table \ref{tab:appendix_hotpot} shows examples on HotpotQA, and Table \ref{tab:appendix_fever} shows examples on FEVER. We should note that QFE does not necessarily extract the sentence with the highest probability score at any step because QFE determines the evidence by using the beam search algorithm.
    
    Three or four correct evidence sentences are extracted in the first and second examples in Table \ref{tab:appendix_hotpot}. The third example is a typical mistake of QFE; QFE extracts too few evidence sentences. In the fourth example, QFE extracts too many evidence sentences. The fifth and sixth questions are typical yes/no questions in HotpotQA. However, like other QA models, our model makes mistakes in answering such easy questions.
    
    One or two evidence sentences are extracted correctly in the first, second, and third examples in Table \ref{tab:appendix_fever}. In FEVER, most claims requiring two evidence sentences can be verified by either of two correct evidence sentences, like in the second example. However, there are some claims that require both evidence sentences, like the third example. The fourth example is a typical mistake of QFE; QFE extracts too few evidence sentences. In the fifth and sixth example, the answers of the questions are `Not Enough Info'. QFE unfortunately extracts evidence when the QA model predicts another label.

	\begin{table}[t]
	\begin{center}
			\scalebox{0.9}{
			\begin{tabular}{ccc}\hline
					 & HotpotQA & FEVER  \\ \hline
					 size of word vectors: $d_w$ & 400 & 435 \\
					 width of RNN: $d_c$ & 150 & 150 \\
					 dropout keep ratio & 0.8 & 0.8 \\
					 batch size & 72 & 96 \\
					 learning rate & 0.001 & 0.001 \\
					 beam size & 5 & 5 \\ \hline
					\end{tabular}}
		\caption{Hyper Parameters.}
		\label{tab:hypara}
	\end{center}
	\end{table}
	
	\begin{table*}[t]
		\begin{center}
			\scalebox{0.8}{
			\begin{tabular}{ccll}\hline
					\multicolumn{4}{l}{$Q$: What plant has about 40 species native to Asia , Manglietia or Abronia?} \\
					\multicolumn{4}{l}{$A_T$: Manglietia, \quad $\hat{A}_T$: Manglietia } \\ \hline
					gold & predicted & probability[\%] & text \\ \hline
					\checkmark & 1 & $\textbf{85.4}$ & 
					\parbox{25em}{\strut{}Abronia ... is a genus of about 20 species of ....
			        \strut} \\ 
					\checkmark & 2 & $14.6 \to \textbf{54.5}$ & 
					\parbox{25em}{\strut{}Manglietia is a genus of flowering plants in the family Magnoliaceae.
			        \strut} \\ 
			        \checkmark & 3 & $~~0.0 \to 45.1 \to \textbf{61.4}$ & 
					\parbox{25em}{\strut{}There are about 40 species native to Asia. \strut} \\ 
					 & 4 & $~~0.0 \to ~~0.0 \to 37.2 \to \textbf{99.2}$  & 
					\parbox{25em}{\strut{}EOE sentence \strut} \\ \hline \hline 
			        
			        \multicolumn{4}{l}{$Q$:Ricky Martin's concert tour in 1999 featured an American heavy metal band formed in what year?} \\
					\multicolumn{4}{l}{$A_T$:1991, \quad $\hat{A}_T$: 1991 } \\ \hline
					gold & predicted & probability[\%] & text \\ \hline
					\checkmark & 1 & $\textbf{100.0}$ & 
					\parbox{25em}{\strut{}Formed on October 12, 1991, the group was founded by vocalist/guitarist Robb Flynn and bassist Adam Duce.
			        \strut} \\ 
					\checkmark & 2 & $~~0.0 \to \textbf{98.8}$ & 
					\parbox{25em}{\strut{}Other bands that were featured included Machine Head, Slipknot, and Amen.
			        \strut} \\ 
			        \checkmark & 3 & $~~0.0 \to ~~0.0 \to \textbf{97.7}$ & 
					\parbox{25em}{\strut{}Machine Head is an American heavy metal band from Oakland, California. \strut} \\ 
					\checkmark & 4 & $~~0.0 \to ~~1.1 \to ~~1.4 \to \textbf{97.3}$ & 
					\parbox{25em}{\strut{}Livin La Vida Loco ... by Ricky Martin, was a concert tour in 1999. \strut} \\ 
					 & 5 & $~~0.0 \to ~~0.0 \to ~~0.9 \to ~~2.0 \to \textbf{97.0}$  & 
					\parbox{25em}{\strut{}EOE sentence \strut} \\ \hline \hline 
			        
					\multicolumn{4}{l}{$Q$: Where is the singer of "B Boy" raised?} \\
					\multicolumn{4}{l}{$A_T$: Philadelphia, \quad $\hat{A}_T$: Philadelphia } \\ \hline
					gold & predicted & probability[\%] & text \\ \hline
					\checkmark & 1 & \textbf{100.0} & 
					\parbox{25em}{\strut{}Raised in Philadelphia, he embarked ....
			        \strut} \\ 
					\checkmark & 2 & $~~0.0 \to \textbf{100.0}$ & 
					\parbox{25em}{\strut{}"B Boy" is a song by American hip hop recording artist Meek Mill.
			        \strut} \\ 
			         & 3 & $~~0.0 \to ~~0.0 \to \textbf{79.0}$ & 
					\parbox{25em}{\strut{}EOE sentence \strut} \\ 
					 \checkmark & --- & $~~0.0 \to ~~0.0 \to 20.8$& 
					\parbox{25em}{\strut{}Robert Rihmeek Williams ... known by his stage name, Meek Mill, ....
			        \strut} \\ \hline \hline 
			        
					\multicolumn{4}{l}{$Q$: Which comic series involves characters such as Nick Fury and Baron von Strucker?} \\
					\multicolumn{4}{l}{$A_T$: Marvel, \quad $\hat{A}_T$: Sgt. Fury } \\ \hline
					gold & predicted & probability[\%] & text \\ \hline
					\checkmark & 1 & \textbf{70.7} & 
					\parbox{25em}{\strut{}Andrea von Strucker ... characters appearing in American comic books published by Marvel Comics.
			        \strut} \\ 
					 & 2 & $~~1.7 \to \textbf{41.6}$ & 
					\parbox{25em}{\strut{}It is the first series to feature Nick Fury Jr. as its main character.
			        \strut} \\ 
			         & 3 & $17.3 \to 38.2 \to \textbf{31.6}$ & 
					\parbox{25em}{\strut{}Nick Fury is a 2017 ongoing comic book series published by Marvel Comics. \strut} \\ 
					 & 4 & $~~0.0 \to ~~0.0 \to 41.6 \to \textbf{92.0}$ & 
					\parbox{25em}{\strut{}EOE sentence \strut} \\ 
					 \checkmark & --- & $~~0.0 \to ~~0.0 \to ~~0.0 \to ~~0.0$& 
					\parbox{25em}{\strut{}Nick Fury: ... the Marvel Comics character Nick Fury.
			        \strut} \\ \hline \hline 
			        
			        \multicolumn{4}{l}{$Q$: Are both "Cooking Light" and "Vibe" magazines?} \\
					\multicolumn{4}{l}{$A_T$: yes, \quad $\hat{A}_T$: yes } \\ \hline
					gold & predicted & probability[\%] & text \\ \hline
					\checkmark & 1 & $\textbf{89.0}$ & 
					\parbox{25em}{\strut{}Cooking Light is an American monthly food and lifestyle magazine founded in 1987.
			        \strut} \\ 
					\checkmark & 2 & $11.0 \to \textbf{97.4}$ & 
					\parbox{25em}{\strut{}Vibe is an American music and entertainment magazine founded by producer Quincy Jones.
			        \strut} \\ 
			         & 3 & $~~0.0 \to ~~0.0 \to \textbf{95.4}$  & 
					\parbox{25em}{\strut{}EOE sentence \strut} \\ \hline \hline 
					
					\multicolumn{4}{l}{$Q$: Are Robert Philibosian and David Ignatius both politicians?} \\
					\multicolumn{4}{l}{$A_T$: no, \quad $\hat{A}_T$: yes } \\ \hline
					gold & predicted & probability[\%] & text \\ \hline
					\checkmark & 1 & $\textbf{100.0}$ & 
					\parbox{25em}{\strut{}Robert Harry Philibosian (born 1940) is an American politician.
			        \strut} \\ 
					\checkmark & 2 & $~~0.0 \to \textbf{98.7}$ & 
					\parbox{25em}{\strut{}David R. Ignatius (May 26, 1950), is an American journalist and novelist.
			        \strut} \\ 
			         & 3 & $~~0.0 \to ~~0.0 \to \textbf{97.4}$  & 
					\parbox{25em}{\strut{}EOE sentence \strut} \\ \hline \hline 
			\end{tabular}}
		\end{center}
		
		\caption{Outputs of QFE on HotpotQA. The sentences are extracted in the order shown in the predicted column. The extraction scores of the sentences at each step are in the probability column.}
		\label{tab:appendix_hotpot}
	\end{table*}

	\begin{table*}[t]
		\begin{center}
			\scalebox{0.85}{
			\begin{tabular}{ccll}\hline
					\multicolumn{4}{l}{$Q$: Fox 2000 Pictures released the film Soul Food.
					\quad $A_T$: Supports \quad $\hat{A}_T$: Supports} \\ \hline
					gold & predicted & probability[\%] & text \\ \hline
					\checkmark & 1 & \textbf{98.0} & 
					\parbox{30em}{\strut{}Soul Food is a 1997 American comedy-drama film ... and released by Fox 2000 Pictures.
			        \strut} \\ 
                     & 2 & $~~0.0 \to \textbf{75.9}$ & 
					\parbox{30em}{\strut{}EOE sentence \strut} \\ \hline \hline 
					
					\multicolumn{4}{l}{$Q$: Terry Crews was a football player.
					\quad $A_T$: Supports \quad $\hat{A}_T$: Supports} \\ \hline
					gold & predicted & probability[\%] & text \\ \hline
					\checkmark & 1 & \textbf{96.0} & 
					\parbox{30em}{\strut{}Terry Alan Crews ... is an American actor , artist , and former American football player.
			        \strut} \\ 
					\checkmark & 2 & $~~3.8 \to \textbf{56.4}$ & 
					\parbox{30em}{\strut{}In football , Crews played as ....
			        \strut} \\ 
			         & 3 & $~~0.0 \to 41.8 \to \textbf{86.4}$ & 
					\parbox{30em}{\strut{}EOE sentence \strut} \\ \hline \hline 
					
					\multicolumn{4}{l}{$Q$: Jack Falahee is an actor and he is unknown.
					\quad $A_T$: Refutes \quad $\hat{A}_T$: Refutes} \\ \hline
					gold & predicted & probability[\%] & text \\ \hline
					\checkmark & 1 & \textbf{95.1} & 
					\parbox{30em}{\strut{}Jack Ryan Falahee (born February 20 , 1989) is an American actor.
			        \strut} \\ 
					\checkmark & 2 & $~~4.9 \to \textbf{67.1}$ & 
					\parbox{30em}{\strut{}He is known for his role as Connor Walsh on ....
			        \strut} \\ 
			         & 3 & $~~0.0 \to 32.9 \to \textbf{100.0}$ & 
					\parbox{30em}{\strut{}EOE sentence \strut} \\ \hline \hline 
					
					\multicolumn{4}{l}{$Q$: Same Old Love is disassociated from Selena Gomez.
					\quad $A_T$: Refutes \quad $\hat{A}_T$: Refutes} \\ \hline
					gold & predicted & probability[\%] & text \\ \hline
					\checkmark & 1 & \textbf{75.0} & 
					\parbox{30em}{\strut{}``Same Old Love'' is a song by American singer Selena Gomez ....
			        \strut} \\ 
					 & 2 & $~~0.0 \to \textbf{69.8}$ & 
					\parbox{30em}{\strut{}EOE sentence
			        \strut} \\ 
			         \checkmark & --  & $~~0.2 \to ~~0.5$ & 
					\parbox{30em}{\strut{}Gomez promoted ``Same Old Love'' .... \strut} \\ \hline\hline 
					
					\multicolumn{4}{l}{$Q$: Annette Badland was in the 2015 NBA Finals
					\quad $A_T$: Not Enough Info \quad $\hat{A}_T$: Not Enough Info} \\ \hline
					gold & predicted & probability[\%] & text \\ \hline
					 & 1 & \textbf{98.6} & 
					\parbox{30em}{\strut{}EOE sentence \strut} \\ \hline\hline 
					
					\multicolumn{4}{l}{$Q$: Billboard Dad is a genre of music.
					\quad $A_T$: Not Enough Info \quad $\hat{A}_T$: Refutes} \\ \hline
					gold & predicted & probability[\%] & text \\ \hline
					 & 1 & \textbf{98.4} & 
					\parbox{30em}{\strut{}Billboard Dad (film) is a 1998 American direct-to-video comedy film .... \strut} \\
					& -- & $~~0.00 \to \textbf{83.5}$ & 
					\parbox{30em}{\strut{}EOE sentence \strut} \\ \hline 
			\end{tabular}}
		\end{center}
		\caption{Outputs of QFE (single model) on FEVER. The sentences are extracted in the order shown in the predicted column. The extraction scores of the sentences at each step are in the probability column.}
		\label{tab:appendix_fever}
	\end{table*}
\end{document}